\documentclass[conference]{IEEEtran}
\IEEEoverridecommandlockouts
\usepackage{hyperref}
\usepackage{cite}
\usepackage{amsmath,amssymb,amsfonts}
\usepackage{algorithmic}
\usepackage{graphicx}
\usepackage{textcomp}
\usepackage[numbers,sort&compress]{natbib}
\usepackage[table,xcdraw]{xcolor}
\def\BibTeX{{\rm B\kern-.05em{\sc i\kern-.025em b}\kern-.08em
    T\kern-.1667em\lower.7ex\hbox{E}\kern-.125emX}}
\begin{document}

\title{RSAttAE: An Information-Aware Attention-based Autoencoder Recommender System
}

\author{\IEEEauthorblockN{1\textsuperscript{st} Amirhossein Dadashzadeh Taromi}
\IEEEauthorblockA{\textit{Department of CS \& IT} \\
\textit{Institute for Advanced Studies}
\\\textit{in Basic Sciences (IASBS)}\\
Zanjan, Iran \\
dadashzadeh@iasbs.ac.ir}
\and

\IEEEauthorblockN{2\textsuperscript{nd} Sina Heydari}
\IEEEauthorblockA{\textit{Department of CS \& IT} \\
\textit{Institute for Advanced Studies}
\\\textit{in Basic Sciences (IASBS)}\\
Zanjan, Iran \\
sinaheydari@iasbs.ac.ir}
\and

\IEEEauthorblockN{3\textsuperscript{rd} Mohsen Hooshmand}
\IEEEauthorblockA{\textit{Department of CS \& IT} \\
\textit{Institute for Advanced Studies}
\\\textit{in Basic Sciences (IASBS)}\\
Zanjan, Iran \\
mohsen.hooshmand@iasbs.ac.ir}
\and

\IEEEauthorblockN{4\textsuperscript{th} Majid Ramezani}
\IEEEauthorblockA{\textit{Department of CS \& IT} \\
\textit{Institute for Advanced Studies}
\\\textit{in Basic Sciences (IASBS)}\\
Zanjan, Iran \\
ramezani@iasbs.ac.ir}
}

\maketitle

\begin{abstract}
Recommender systems play a crucial role in modern life, including information retrieval, the pharmaceutical industry, retail, and entertainment. The entertainment sector, in particular, attracts significant attention and generates substantial profits. This work proposes a new method for predicting unknown user-movie ratings to enhance customer satisfaction. To achieve this, we utilize the MovieLens 100K dataset. Our approach introduces an attention-based autoencoder to create meaningful representations and the XGBoost method for rating predictions. The results demonstrate that our proposal outperforms most of the existing state-of-the-art methods.\\
Availability:~\href{https://github.com/ComputationIASBS/RecommSys}{github.com/ComputationIASBS/RecommSys}
\end{abstract}


\section{Introduction}\label{sec:intro}
In today's digital age, the Internet plays a crucial role in helping people find the information and services they need. Users often seek guidance in their searches, while companies strive to present their products and services to potential customers. This is where the concept of ``recommender systems'' comes into play. The recommender systems aim to understand user interests and suggest new services or products based on that understanding~\cite{li24}. Significant effort has been dedicated to improving the performance of recommendation systems, utilizing various computational and AI methods~\cite{rssurvey22}. However, there is still potential for further enhancement in these systems. We propose a \textbf{R}ecommender \textbf{S}ystem and introduce an \textbf{At}tention-based \textbf{A}uto \textbf{E}ncoder (RSAttAE). 
The proposed framework involves two RSAttAE modules that generate new user and movie representations using the rating matrix while incorporating entities' features. These new representations are then fed into a Machine Learning (ML) algorithm to predict the ratings. We utilize the MovieLens dataset to evaluate the performance of the RSAttAE module. The results demonstrate the effectiveness and robustness of our proposed method in rating prediction or matrix completion tasks.

Section~\ref{sec:rel} discusses related work, particularly studies that have achieved the best results in the recommender systems literature. Section~\ref{sec:data} describes the dataset used in this study. Section~\ref{sec:res} presents the results and performance of our proposed method. Finally, Section~\ref{sec:conc} concludes the paper. 

\section{Realted work}\label{sec:rel}
The recommender system methodologies can be categorized into three main classes: \textit{Content-based filtering}, \textit{Collaborative filtering}, and \textit{Hybrid filtering} methods. While content-based methods leverage the entities' features and user's history to recommend items to the active user, collaborative filtering~\cite{pfeae18, graem20, glocal21} techniques mainly focus on stating similar entities based on their interactions and suggest similar items to similar users. Both of each have their advantages and limitations, though a hybrid view~\cite{graphrec19, mggat20, wmlff23} tries to benefit from the advantages of each method and simultaneously compensate for the limitations~\cite{rssurvey22}. This section reviews some influential studies in the literature. Most state-of-the-art approaches belong to the latter categories. Therefore, we introduce related studies in these two categories.
\subsection{Collaborative Filtering methods:}

Hartford et al.~\cite{pfeae18} discussed the issue of the interchangeability of users and movies. In other words, most of the proposed methods are sensitive to the order of users and movies. Instead, the authors vectorized the rating matrix. Their goal was to keep the learning methods permutation-invariant. Therefore, the input matrix is converted to a vector representation. In other words, they tried to convert the representation to maintain the set properties of the input data. They proposed two learning methods, i.e., self-supervised using a fully connected network, and an autoencoder. They used an autoencoder to complete the matrix of the user-item ratings.  

Moreover, Strahl et al.~\cite{graem20} proposed a graph-based approach to include side information for rating predictions. In other words, They discussed the importance of demographic information of users and movies' side information. They create graphs for the representations of this mentioned data. However, the generated graph from the side information may have inconsistencies with the latent feature of rating matrix embeddings. Therefore, the authors used a graph-based lasso method and utilized expectation maximization to remove irrelevant edges from the side information graphs. 

In 2021, Han et al.~\cite{glocal21} proposed an autoencoder-based approach for prediction ratings in recommender systems. They input the rating matrix into an autoencoder, using a reconstruction loss function. In their method, the decoder component was left unchanged, while the encoder weights were multiplied by a radial basis function (RBF) kernel, serving as a local kernel. They then defined a $(3 \times 3)$ global kernel for each item, multiplying the corresponding kernel by each item's average rating. In the next step, they created a global kernel by summing the global kernels of all items. The rating matrix was then convolved with this global kernel, and the result was fed into the trained autoencoder. The final reconstructed matrix was considered the prediction for unknown user-item interactions. 

\subsection{Hybrid Filtering methods:}

Rashed et al.~\cite{graphrec19} discussed two challenges related to embeddings: the scarcity of user information due to privacy concerns and the computationally intensive methods for co-embedding additional user and item features. To tackle these challenges in recommender systems, they proposed a nonlinear graph-based method known as the attribute-aware approach. For user-item interactions, the authors concatenated three distinct pieces of information. The first part originates from the rating matrix, the second from profile features, and the third from the concatenation of each user node's degree with the Laplacian of its corresponding feature vector. The user and item embeddings were fed into a fully connected layer to reduce the prediction model's data dimensions and complexity while applying nonlinearity to the final feature vector. For rating prediction, an inner product is applied to the respective final feature vectors of users and items. 

Later, Leng et al.~\cite{mggat20} focused on smoothing the embedding by proposing a graph-based method that utilizes the graph attention network (GAT) on user and item graphs. In the first step, their method generated user and item graphs. Each graph was concurrently fed into a fully connected layer. The output was then input into a GAT, which aimed to smooth the new embedding using the close neighbors of each node (item and user). The smoothed feature vectors were subsequently fed into another fully connected layer. In the final step, they defined independent global smoothers by employing the Laplacian matrices and graph regularization terms. The final embeddings are the sum of GAT embeddings and the Laplacian independent features. They utilized an attention mechanism based on the rating matrix and the generated embedding.

Also, the study of Rodriguez et al.~\cite{wmlff23} is noteworthy where they defined a specific neural network architecture to capture low and high levels of feature abstractions to predict the ratings robustly. Each layer of their proposed architecture contains one fully connected layer followed by a noise adder. They used several of the mentioned layers to extract different levels of abstraction. The prediction result for each user-movie rating is equal to a weighted sum of the inner products of each level of abstraction.  

% [ ] TODO:: Add a sudo-code 
% [X] Plots
% [X] Introduction
% [ ] Metrics

\section{Dataset}\label{sec:data}
%\textcolor{red}{MovieLens:} Harper and Knostan~\cite{movielens15} proposed MovileLens datasets 

\begin{figure*}
    \centering
    \includegraphics[width=0.98\textwidth]{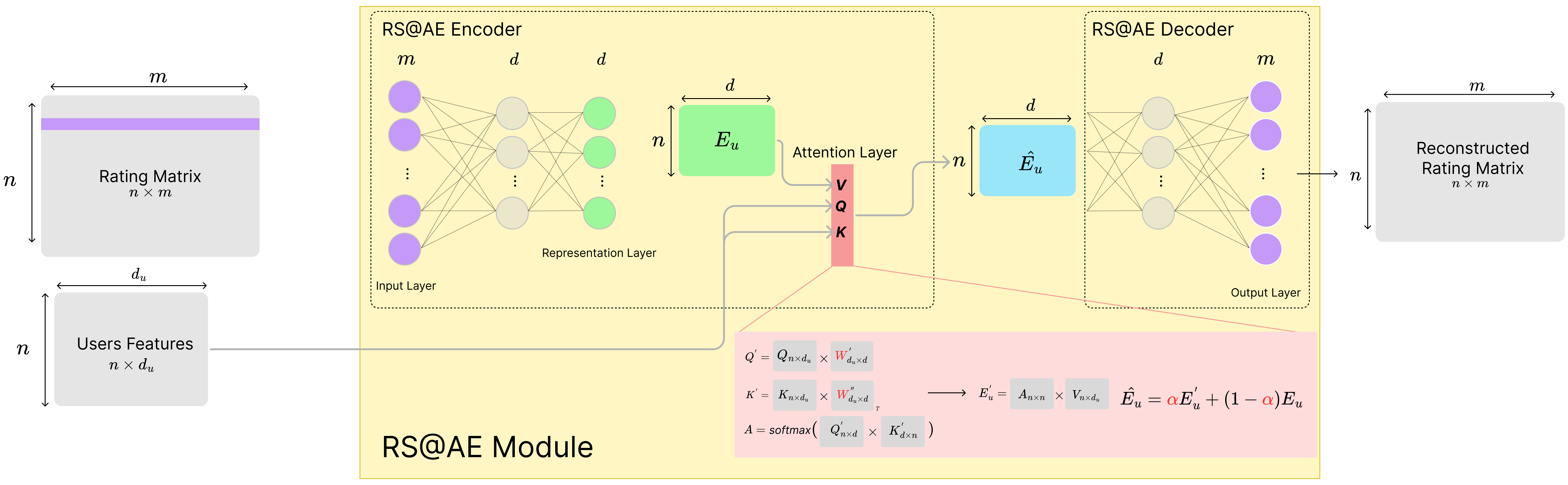}
    \caption{The RSAttAE module architecture consists of attention-based encoder and decoder modules.}
    \label{fig:sr@ae-architecture}
\end{figure*}

\begin{figure}
    \centering
    \includegraphics[width=0.85\linewidth]{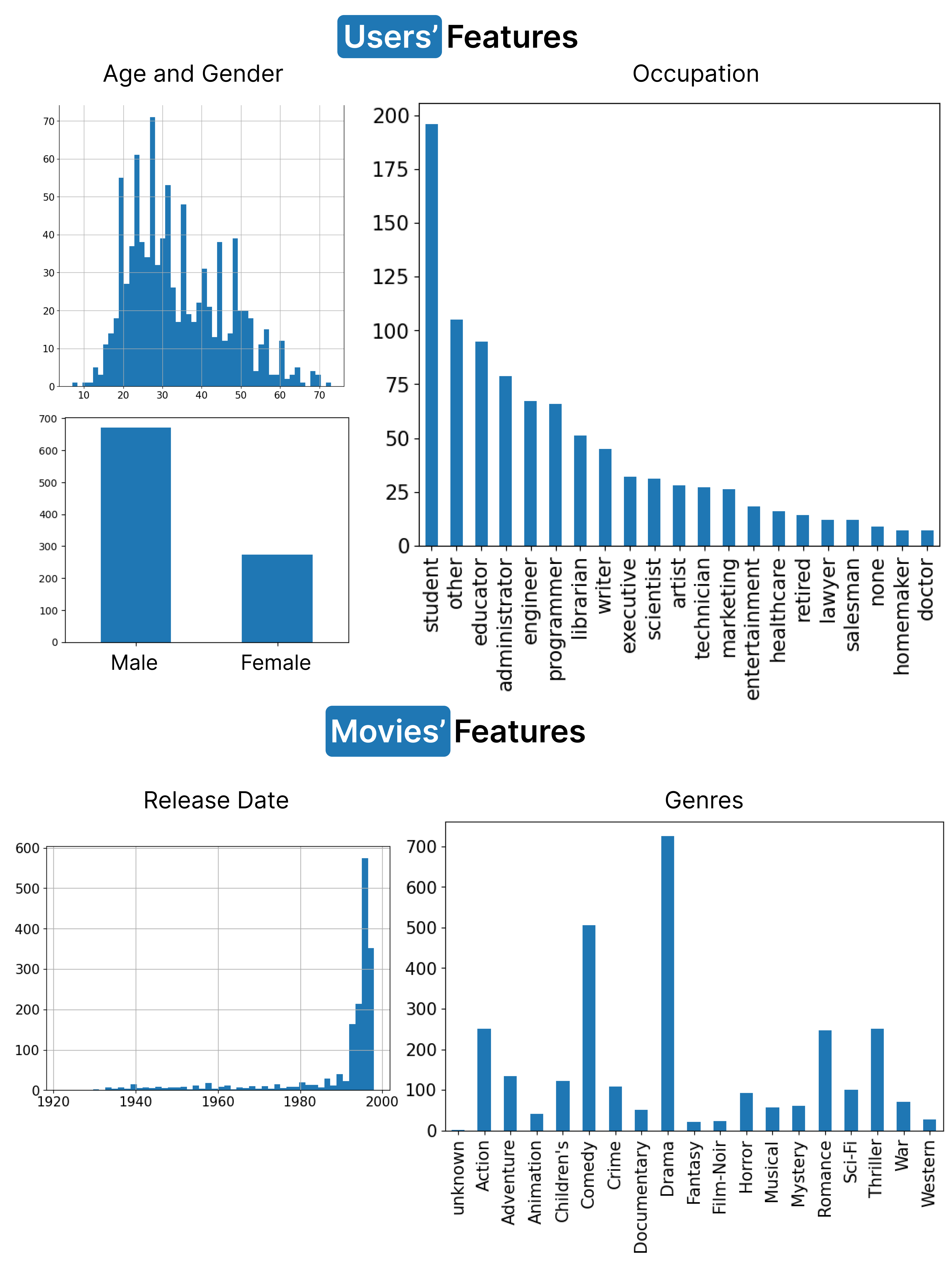}
    \caption{Features distribution of MovieLens 100k}
    \label{fig:features_stats}
\end{figure}

University of Minnesota researchers started developing the MovieLens platform in 1997 which was used to suggest movies to users based on their interactions and ratings. Since then they also collected datasets in various sizes with 100K, 1M, 10M, and 20M user-movie interactions. Alongside the interactions, MovieLens also involves demographic attributes such as age, gender, user occupation, release date, IMDB URL, and genres for movies that provide auxiliary information. In recommender systems literature, ML 100k and ML 1M are widely used as standard benchmark datasets to assess the performance of proposed methods in the context of matrix completion or rating prediction. Table~\ref{table:movielens-stats} briefly shows the statistics related to the datasets collected by the MovieLens group. In the current research, we mainly focus on the ML 100K dataset to assess and compare the performance of our proposed method to other algorithms in the literature. Furthermore, the user and movie feature distributions of MovieLens 100k are illustrated in Fig~\ref{fig:features_stats} for further acquaintance with the dataset.

\begin{table}[]
\centering
\caption{MovieLens datasets' statistics \cite{movielens15}}
\label{table:movielens-stats}
\begin{tabular}{llllll}
Name    & Rating Scale & Users   & Movies & Ratings    & Density \\
\hline
ML 100K & 1-5          & 943     & 1,682  & 100,000    & 6.30\%  \\
ML 1M   & 1-5          & 6,040   & 3,706  & 1,000,209  & 4.47\%  \\
ML 10M  & 0.5-5        & 69,878  & 10,681 & 10,000,054 & 1.34\%  \\
ML 20M  & 0.5-5        & 138,493 & 27,278 & 20,000,263 & 0.54\% 
\end{tabular}
\end{table}
% TODO[X]:: Fonts of picture is small

%\textcolor{red}{Data Statistics:}

\section{Proposed method}\label{sec:prop}

The proposed method consists of three main phases: preprocessing, encoding, and decoding. In the preprocessing phase, we focus on categorizing features. The encoding phase employs the proposed module RSAttAE to create meaningful and informative representations of users and movies. Finally, the decoding phase is dedicated to predicting ratings.

\subsection{Preprocessing}

In the first step, we begin by preprocessing and selecting features. For users, we consider the features of age, gender, and occupation. For movies, we use the release year and genres. Next, we categorize the numerical features—age and release year—to reduce noise and ensure that the feature vectors for both users and movies are compatible. The age bins are set at (0, 12, 18, 30, 50, 90) and the release year bins are (1920, 1940, 1960, 1980, 2000). Finally, we generate one-hot encoded vectors for each feature, resulting in feature matrices $ F_u \in {R}^{n \times d_u}$ for users and $F_m \in {R}^{m \times d_m}$ for movies, where $n$ and $m$ represent the number of users and movies, and $d_u$ and $d_m$ denote the number of features for users and movies, respectively.

\subsection{Framework}
The primary focus of RSAttAE consists of two key phases:

\begin{enumerate}

    \item \textbf{Training RSAttAE:} We train two RSAttAEs to reconstruct both the user-rating matrix and its transpose. This process allows us to learn embeddings for users and movies, respectively. Moreover, the RSAttAE incorporates attention mechanism to leverage entities' features. 

    \item \textbf{Applying Supervised Learning:} To enhance the efficiency of the collaborative filtering task, we implement a supervised learning approach using the \textit{XGBoost} model to predict ratings.
\end{enumerate}
% TODO [] find citation
% TODO [] remove, from equations

In literature, methods that use autoencoders to learn non-linear latent features have consistently outperformed those based solely on linear Matrix Factorization (MF) or Singular Value Decomposition (SVD). Additionally, incorporating user and movie features to generate embeddings is crucial, as it not only enhances rating predictions but also helps address the cold start problem, which remains a significant challenge in recommender systems~\cite{graphrec19}. Some studies, such as GraphRec~\cite{graphrec19} and MG-GAT~\cite{mggat20}, employ graph-based techniques to co-embed the rating matrix and feature information. In the current study, we modify the standard AE architecture to implicitly incorporate entities' features by using an attention mechanism akin to that proposed by Vaswani et al. in their work on Transformers~\cite{attention2017}.

\begin{figure*}
    \centering
\includegraphics[width=0.7\textwidth]{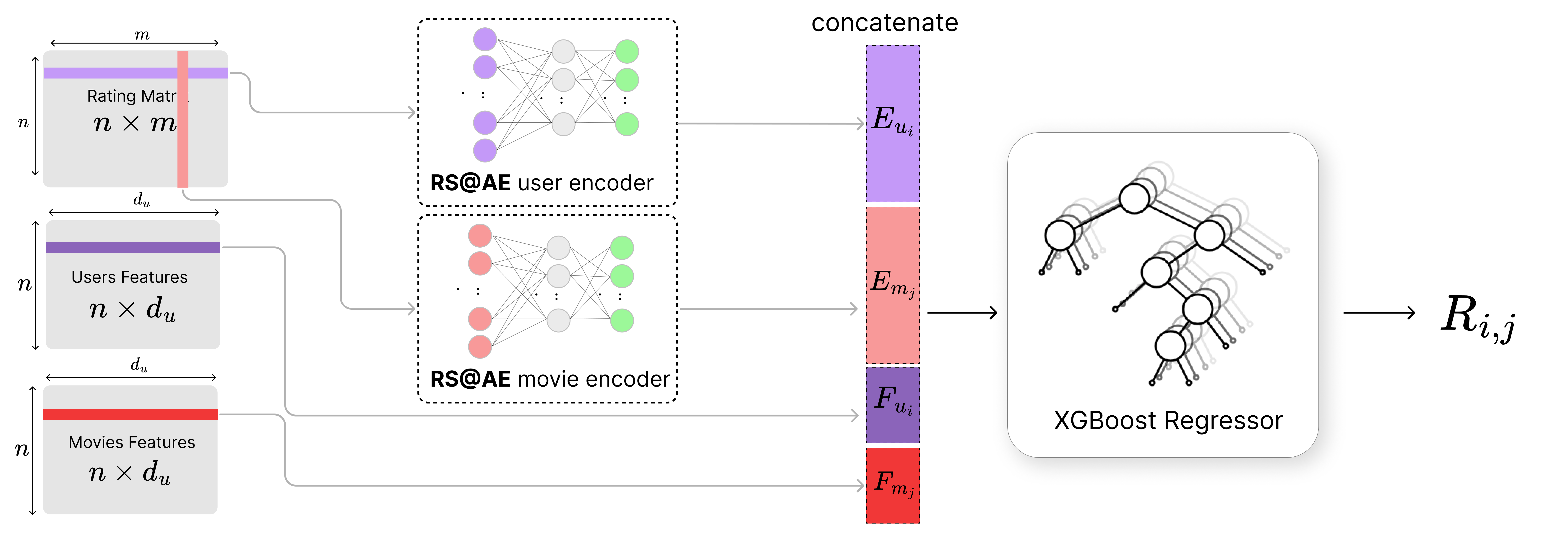}
    \caption{The rating prediction procedure using the XGBoost algorithm and the trained RSAttAE encoders.}
    \label{fig:xgboost-training}
\end{figure*}

\subsection{Attention-based Autoencoder}
The overall architecture of Attention-based Autoencoder or SR@AE module is depicted in Fig~\ref{fig:sr@ae-architecture}, and we review its mechanisms in more detail. Considering $R \in {R}^{n \times m}$ as the rating matrix, we utilize a two-layer fully connected network to map each user into the latent space, resulting in $E_u \in {R}^{n \times d}$. The following equation defines the relationship:

\begin{equation}
    E_u = \sigma(\sigma(R W + b) \hat{W} + \hat{b}),
\end{equation}

where $W \in {R}^{m \times d}$ and $\hat{W} \in {R}^{d \times d}$ are the weight matrices of the first and second fully connected layers, respectively. The vectors $b$ and $\hat{b} \in {R}^{d}$ represent the biases. The symbol $\sigma$ denotes a nonlinear activation function, the Leaky ReLU in this context. The choice of Leaky ReLU helps to address issues associated with the dying gradient problem of ReLU and the vanishing gradient problem observed in sigmoid functions.

While the vanilla Autoencoder (AE)~\cite{autoencoder12} directly utilizes the generated embedding $E_u$ to reconstruct the original data, RSAttAE module employs an attention mechanism to update $E_u$. In other words, RSAttAE generates a new embedding $\hat{E}_u$ by convexly combining the latent feature $E_u$ from the AE with an embedding from the attention mechanism considering the users' features. 

The attention mechanism is akin to the cross-attention layer found in the Transformer architecture~\cite{attention2017}. The following equations illustrate the formulation of the enhanced embeddings. 
% TODO [] dimensions with small characters

\begin{equation}
    \label{eq:attention}
    \begin{split}
        & Q^{'} = QW^{'} \\
        & K^{'} = KW^{''} \\
        & V = E_u \\
        & A = softmax(Q^{'}K'^{T}) \\
        & \hat{E_u} = LayerNorm(\alpha AV + (1-\alpha) E_u)
    \end{split}
\end{equation}

The matrices $W^{'}$ and $\hat{W}^{''}$ in Eq~\eqref{eq:attention} belong to $R^{d_u \times d}$ and serve as projection matrices that map features into a $d$-dimensional space. After performing row-wise normalization of the matrices $Q^{'}$ and $K^{'}$, we compute the similarity matrix by multiplying $Q^{'}$ and $K^{'T}$. Subsequently, we apply row-wise softmax to obtain the attention weights. The embeddings from the attention layer are then computed by multiplying $A$ and $V$ matrices. The parameter $\alpha$ is a learnable parameter that determines the influence of the attention layer. In this context, $Q$, $K$, and $V$ correspond to the Query, Key, and Value in the attention mechanism. Then, another two-layer fully connected network is utilized to reconstruct the user-item rating matrix using $\hat{E}_u$ representation. Interestingly, the $\alpha$ parameter after training converges to ($0.5\leq \alpha \leq 0.7$) which shows the importance of the attention layer.

\begin{equation}
    \hat{R} = \sigma(\hat{W}\sigma(W\hat{E_u} + b)+\hat{b})
\end{equation}

The loss function is defined as the \textit{masked} root mean squared error (M-RMSE) which is declared as:

\begin{equation}
    \text{M-RMSE }= \sqrt{\frac{\sum_{r_{i,j} \in R;r_{i,j} \neq 0}{(r_{i,j} - \hat{r}_{i,j}})^2}{N}}
\end{equation}

Where $N$ is the number of all non-zero elements in the rating matrix.

\subsection{Training XGBoost model}

After training RSAttAE modules for users and movies, we constructed a supervised learning setting similar to COFILS~\cite{cofils17}. If $R_{i,j}$ is a non-zero element in the rating matrix, then the dataset contains input feature $x = concat[E_{u_i}, E_{m_j}, F_{u_i}, F_{m_j}]$ and the target label $R_{i,j}$. The mentioned procedure is illustrated in Fig~\ref{fig:xgboost-training}. Similar to the strategy used to train, validate, and test the RSAttAE encoders, the constructed supervised dataset also consisted of three subsets of the train, validation(10 percent of `u1.base`), and test.

\section{Results}\label{sec:res}

\subsection{Embedding Dimension($d$)}
\begin{figure}
    \centering
    \includegraphics[width=\linewidth]{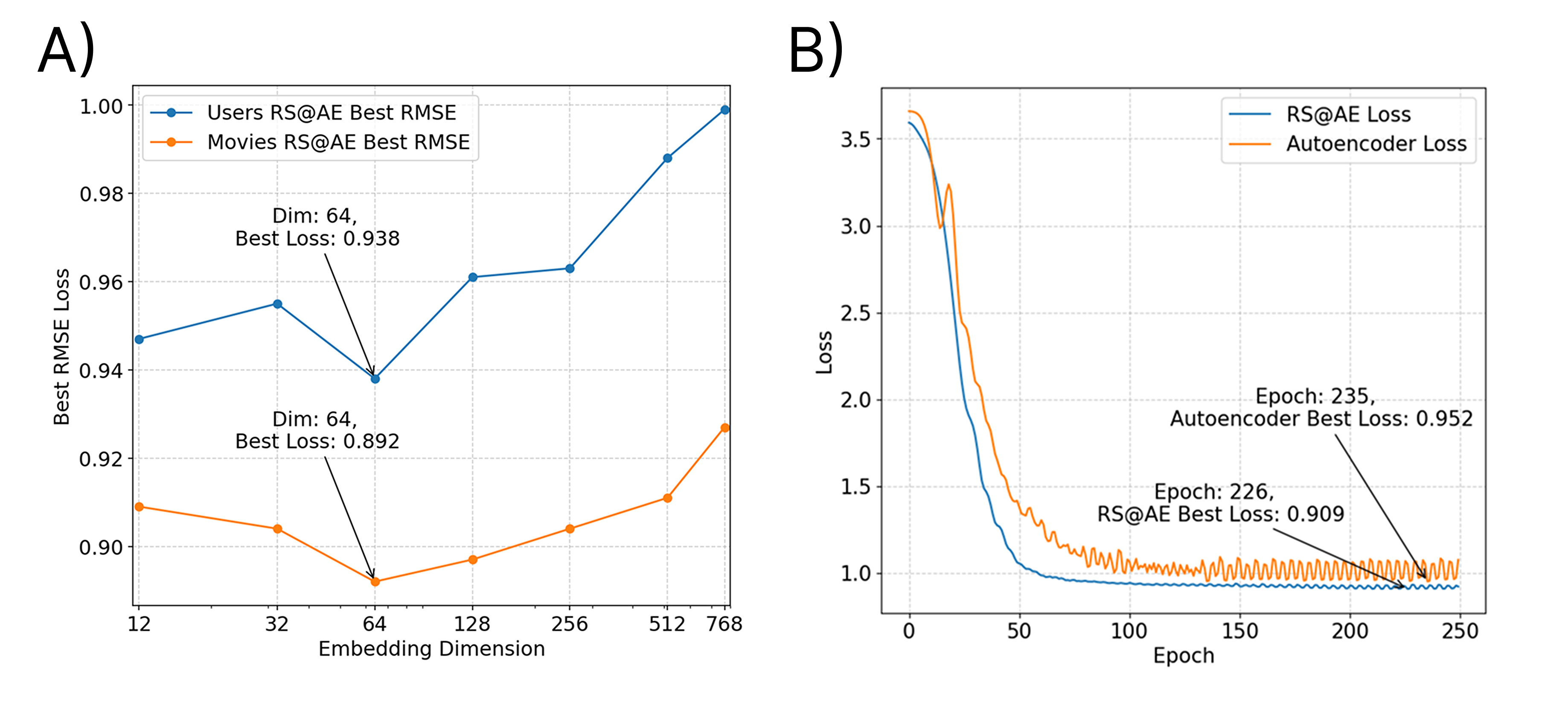}
    \caption{A: Validation loss vs Embedding dimension, which indicates $d=64$ as the best embedding dimension. B: Validation loss vs Epochs for both vanilla Autoencoder and RSAttAE demonstrates the superiority of RSAttAE in convergence.}
    \label{fig:dimension_effect}
\end{figure}

Figure~\ref{fig:dimension_effect}A illustrates the effect of different embedding dimensions on the best RMSE loss for RSAttAE module across users and movies. The results demonstrate that embedding dimension significantly impacts model performance, with the best RMSE loss achieved at a dimension of 64 for both users and movies. Specifically, the lowest reconstruction RMSE loss for users is 0.938, and for movies, it is 0.892. As the embedding dimension increases beyond 64, the RMSE loss begins to rise for both users and movies, indicating diminishing returns and potential overfitting at higher dimensions. This finding highlights the importance of selecting an optimal embedding size for effective representation learning and model performance.

\subsection{XGBoost regressor hyperparameter tuning}
The XGBoost regressor includes several hyperparameters that control model complexity and training behavior which have been included in table~\ref{table:xghyper}. 
The parameter \textit{n\_estimators} specifies the number of boosting rounds, while \textit{max\_depth} controls the depth of each tree, balancing complexity and overfitting. \textit{learning\_rate} determines the step size at each boosting iteration, allowing for a gradual learning process. \textit{subsample} and \textit{colsample\_bytree} introduce randomness by selecting subsets of data and features, respectively, to improve generalization. Regularization parameters, \textit{reg\_lambda} (L2 regularization) and \textit{reg\_alpha} (L1 regularization), prevent overfitting by penalizing large weights and encouraging sparsity in the model.

Several setups were tested and we include four of them, varying only the number of estimators (n\_estimators) while keeping other hyperparameters constant: max\_depth at 9, learning\_rate at 0.01, subsample at 1, colsample\_bytree at 0.2, reg\_lambda at 1, and reg\_alpha at 3. The results show that increasing the number of estimators improves performance, with RMSE decreasing from 0.9006 in Setup 1 (1300 estimators) to 0.8989 in Setup 4 (1900 estimators). While Setup 4 achieves the best RMSE, the improvement diminishes with higher n\_estimators, suggesting that further increases may yield limited benefits or risk overfitting.

\begin{table}[]
\centering
\caption{XGBoost regressor hyperparameter tuning on the validation set. Setup 4 has achieved the lowest RMSE loss.}
\label{table:xghyper}
\begin{tabular}{lcccc}
\textbf{Hyperparameters}   & \multicolumn{1}{l}{\textbf{setup 1}} & \multicolumn{1}{l}{\textbf{setup 2}} & \multicolumn{1}{l}{\textbf{setup 3}} & \multicolumn{1}{l}{\textbf{setup 4}} \\ \hline
\textbf{n\_estimators}     & 1300                                 & 1500                                 & 1700                                 & 1900                                 \\
\textbf{max\_depth}        & 9                                    & 9                                    & 9                                    & 9                                    \\
\textbf{learning\_rate}    & 0.01                                 & 0.01                                 & 0.01                                 & 0.01                                 \\
\textbf{subsample}         & 1                                    & 1                                    & 1                                    & 1                                    \\
\textbf{colsample\_bytree} & 0.2                                  & 0.2                                  & 0.2                                  & 0.2                                  \\
\textbf{reg\_lambda}       & 1                                    & 1                                    & 1                                    & 1                                    \\
\textbf{reg\_alpha}        & 3                                    & 3                                    & 3                                    & 3                                    \\
\textbf{RMSE}              & 0.8705                               & 0.8699                               & 0.8695                               & \textbf{0.8694}                     
\end{tabular}
\end{table}

\subsection{RSAttAE RMSE comparison}
To establish the efficacy of the proposed method, we compare its performance with other methods in the literature, including state-of-the-art methods, based on \textbf{RMSE}, which is the key evaluation metric for recommender systems on the \textbf{ML 100K} dataset on the standard \textbf{u1 Splits}. Additionally, it should be noted that our method uses no extra training data. As shown in Table~\ref{table:results}, the results indicate that our method achieves near state-of-the-art performance in comparison to others. RSAttAE reduces RMSE loss by 0.030 compared to WMLFF~\cite{wmlff23}, 0.019 compared to GRAEM~\cite{graem20}, and 0.022 compared to FactorizedEAE~\cite{feae18}, demonstrating its robustness compared to current methods. Nevertheless, it falls slightly short of GLocal-K~\cite{glocal21}, which currently achieves state-of-the-art performance with the lowest RMSE loss.
% It is worth mentioning that MG-GAT~\cite{mggat20}, which outperforms our method by only a small margin of 0.009 RMSE, has utilized extra training data. % [ ] TODO::Are sure of that?

\begin{table}[]
\centering
\caption{Comparison of \textbf{RSAttAE} with the state-of-the-art recommender system methods}
\label{table:results}
\begin{tabular}{lc}
\textbf{Methods}                                          & \textbf{RMSE (u1 splits)} \\
\hline
\textbf{WMLFF~\cite{wmlff23}}       & 0.928                     \\
\textbf{GLocal-K~\cite{glocal21}}   & 0.888                     \\
\textbf{MG-GAT~\cite{mggat20}}      & 0.890                     \\
\textbf{GRAEM~\cite{graem20}}       & 0.917                    \\
\textbf{GraphRec~\cite{graphrec19}} & 0.904                     \\
\textbf{FactorizedEAE~\cite{feae18}}                & 0.920                     \\
\textbf{RSAttAE}                                            & 0.898           
\end{tabular}
\end{table}

\section{Conclusion}\label{sec:conc}
This work has introduced an attention-based autoencoder (RSAttAE) module to incorporate entities' features in latent feature learning. We explored the effectiveness and robustness of the proposed module. Using the validation set, we tuned the embedding dimension for RSAttAE. Furthermore, by training an XGBoost regressor using grid search for hyperparameter tuning, we shifted the rating prediction task from matrix completion to a supervised setting. Finally, our proposed method showed comparable results and outperformed some of the \textit{SOTA} models in the literature, indicating the SR@AE as a promising direction of research for further studies. 

%\section*{Acknowledgment}

\bibliographystyle{unsrtnat}
\bibliography{ref}

\begin{thebibliography}{13}
\providecommand{\natexlab}[1]{#1}
\providecommand{\url}[1]{\texttt{#1}}
\expandafter\ifx\csname urlstyle\endcsname\relax
  \providecommand{\doi}[1]{doi: #1}\else
  \providecommand{\doi}{doi: \begingroup \urlstyle{rm}\Url}\fi

\bibitem[Li et~al.(2024)Li, Liu, Satapathy, Wang, and Cambria]{li24}
Yang Li, Kangbo Liu, Ranjan Satapathy, Suhang Wang, and Erik Cambria.
\newblock Recent developments in recommender systems: A survey.
\newblock \emph{IEEE Computational Intelligence Magazine}, 19\penalty0 (2):\penalty0 78--95, 2024.

\bibitem[Zangerle and Bauer(2022)]{rssurvey22}
Eva Zangerle and Christine Bauer.
\newblock Evaluating recommender systems: survey and framework.
\newblock \emph{ACM Computing Surveys}, 55\penalty0 (8):\penalty0 1--38, 2022.

\bibitem[Hartford et~al.(2018{\natexlab{a}})Hartford, Graham, Leyton-Brown, and Ravanbakhsh]{pfeae18}
Jason Hartford, Devon Graham, Kevin Leyton-Brown, and Siamak Ravanbakhsh.
\newblock Deep models of interactions across sets.
\newblock In \emph{International Conference on Machine Learning}, pages 1909--1918. PMLR, 2018{\natexlab{a}}.

\bibitem[Strahl et~al.(2020)Strahl, Peltonen, Mamitsuka, and Kaski]{graem20}
Jonathan Strahl, Jaakko Peltonen, Hirsohi Mamitsuka, and Samuel Kaski.
\newblock Scalable probabilistic matrix factorization with graph-based priors.
\newblock In \emph{Proceedings of the AAAI conference on artificial intelligence}, volume~34, pages 5851--5858, 2020.

\bibitem[Han et~al.(2021)Han, Lim, Long, Burgstaller, and Poon]{glocal21}
Soyeon~Caren Han, Taejun Lim, Siqu Long, Bernd Burgstaller, and Josiah Poon.
\newblock Glocal-k: Global and local kernels for recommender systems.
\newblock In \emph{Proceedings of the 30th ACM International Conference on Information \& Knowledge Management}, pages 3063--3067, 2021.

\bibitem[Rashed et~al.(2019)Rashed, Grabocka, and Schmidt-Thieme]{graphrec19}
Ahmed Rashed, Josif Grabocka, and Lars Schmidt-Thieme.
\newblock Attribute-aware non-linear co-embeddings of graph features.
\newblock In \emph{Proceedings of the 13th ACM conference on recommender systems}, pages 314--321, 2019.

\bibitem[Leng et~al.(2020)Leng, Ruiz, Dong, and Pentland]{mggat20}
Yan Leng, Rodrigo Ruiz, Xiaowen Dong, and Alex Pentland.
\newblock Interpretable recommender system with heterogeneous information: A geometric deep learning perspective.
\newblock \emph{SSRN Electron. J}, 10:\penalty0 2411--2430, 2020.

\bibitem[Rodriguez and Tommasel(2023)]{wmlff23}
Juan~Manuel Rodriguez and Antonela Tommasel.
\newblock Weighted multi-level feature factorization for app ads ctr and installation prediction.
\newblock \emph{arXiv preprint arXiv:2308.02568}, 2023.

\bibitem[Harper and Konstan(2015)]{movielens15}
F~Maxwell Harper and Joseph~A Konstan.
\newblock The movielens datasets: History and context.
\newblock \emph{Acm transactions on interactive intelligent systems (tiis)}, 5\penalty0 (4):\penalty0 1--19, 2015.

\bibitem[Vaswani(2017)]{attention2017}
A~Vaswani.
\newblock Attention is all you need.
\newblock \emph{Advances in Neural Information Processing Systems}, 2017.

\bibitem[Baldi(2012)]{autoencoder12}
Pierre Baldi.
\newblock Autoencoders, unsupervised learning, and deep architectures.
\newblock In \emph{Proceedings of ICML workshop on unsupervised and transfer learning}, pages 37--49. JMLR Workshop and Conference Proceedings, 2012.

\bibitem[Barbieri et~al.(2017)Barbieri, Alvim, Braida, and Zimbr{\~a}o]{cofils17}
Julio Barbieri, Leandro~GM Alvim, Filipe Braida, and Geraldo Zimbr{\~a}o.
\newblock Autoencoders and recommender systems: Cofils approach.
\newblock \emph{Expert Systems with Applications}, 89:\penalty0 81--90, 2017.

\bibitem[Hartford et~al.(2018{\natexlab{b}})Hartford, Graham, Leyton-Brown, and Ravanbakhsh]{feae18}
Jason Hartford, Devon Graham, Kevin Leyton-Brown, and Siamak Ravanbakhsh.
\newblock Deep models of interactions across sets.
\newblock In \emph{International Conference on Machine Learning}, pages 1909--1918. PMLR, 2018{\natexlab{b}}.

\end{thebibliography}

\end{document}